\documentclass[letterpaper, 10 pt, conference]{ieeeconf} 

\IEEEoverridecommandlockouts                              

\usepackage{amsmath,amssymb,amsfonts}
\usepackage{algorithm}
\usepackage{algpseudocode}
\usepackage{graphicx}
\usepackage{svg}
\usepackage{subcaption}
\usepackage[font=small]{caption}

\usepackage[
    backend=bibtex,
    style=ieee,
    mincitenames=1,
    maxcitenames=2,
    maxbibnames=6,
    isbn=false,
    url=false,
    doi=false,
    natbib=true]{biblatex}
\usepackage{textcomp}
\usepackage[table]{xcolor}
\usepackage{booktabs}
\usepackage{tabularx}
\usepackage{multicol}
\usepackage[bookmarks=true]{hyperref}
\usepackage[capitalise,nameinlink]{cleveref} 

\usepackage{gensymb}
\usepackage{svg}
\usepackage{todonotes}
\usepackage{multirow}

\usepackage{tikz}
\usetikzlibrary{positioning,arrows.meta,shapes.geometric,calc}

\def\BibTeX{{\rm B\kern-.05em{\sc i\kern-.025em b}\kern-.08em
    T\kern-.1667em\lower.7ex\hbox{E}\kern-.125emX}}

\newcommand\Tstrut{\rule{0pt}{2.6ex}}         

\begin{document}

\title{\LARGE \bf\textit{BayesContact}: Uncertain Pose Estimation via Visuo-Tactile Proposals and Simulation-based Inference}

\author{
Aditya Kamireddypalli$^1$, Mat{\'i}as Mattamala$^{1}$, Jo\~{a}o Moura$^1$, Russell Buchanan$^2$, \\Sethu Vijayakumar$^1$, Subramanian Ramamoorthy$^1$
%
\thanks{$^1$School of Informatics, University of Edinburgh, Scotland. $\texttt{a.kamireddypalli@sms.ed.ac.uk}$}
\thanks{$^2$Department of Mechanical and Mechatronics Engineering, University of Waterloo, Canada.}
\thanks{\noindent A. Kamireddypalli is supported in part by funding in the form of an unrestricted gift from Microsoft Research. S. Ramamoorthy is supported by a UKRI Turing AI World Leading Researcher Fellowship on AI for Person-Centred and Teachable Autonomy (grant EP/Z534833/1).}
\thanks{\noindent For the purpose of open access, the authors have applied a Creative Commons Attribution (CC BY) license to any Author Accepted Manuscript version arising from this submission.}
}


\maketitle

\begin{abstract}
Contact-rich manipulation requires pose estimates that are often more accurate than what depth-only sensing provides.
Existing methods, relying on vision and contact, employ costly offline training procedures that need to be retrained for new environments and geometries.
We propose \textit{BayesContact}, a Simulation-Based Inference framework for visuo-tactile pose estimation in peg-in-hole insertion. BayesContact maintains a particle belief over object pose and fuses depth observations with force/torque-derived contact evidence. 
We employ simulation based forward models to approximate these observation likelihoods.
For each pose hypothesis, a renderer predicts depth measurements and a physics simulator predicts contact outcomes under guarded probing actions; both are scored against real observations to update the belief. The resulting multimodal belief also enables information-gain-based probing for active disambiguation. Across simulated geometries and real-robot experiments, BayesContact improves pose observability and insertion success over vision-only inference by 30\%. 
\end{abstract}


\section{Introduction}
For robots to automate manipulation tasks, they must be capable of reliable
physical interaction with objects. Everyday industrial skills such as key
insertion, cable connection, fastening, and assembly are contact-rich: the robot
must make, maintain, and break contact while respecting tight geometric
constraints. In these settings, small estimation errors in relative pose or local geometry
can lead to missed contacts, jamming, excessive forces, or task failure.
Reliable, contact-rich manipulation depends not only on control, but
also on accurate estimation of the task-relevant scene state.

A central difficulty in these tasks is partial observability. In peg-in-hole
insertion, the robot must reason about the pose of the hole, but this state is
only indirectly observed through vision, proprioception, and contact. Depth
sensing provides useful global information, but suffers from sensor noise,
occlusions, restricted viewpoints, and
ambiguities due to part geometry. The most task-relevant surfaces may be
hidden inside the hole or only revealed through physical interaction; for example, plugging in a USB-A device. As a
result, visually plausible pose hypotheses may not be physically feasible.


\begin{figure}[t]
\captionsetup{font=footnotesize}
\includegraphics{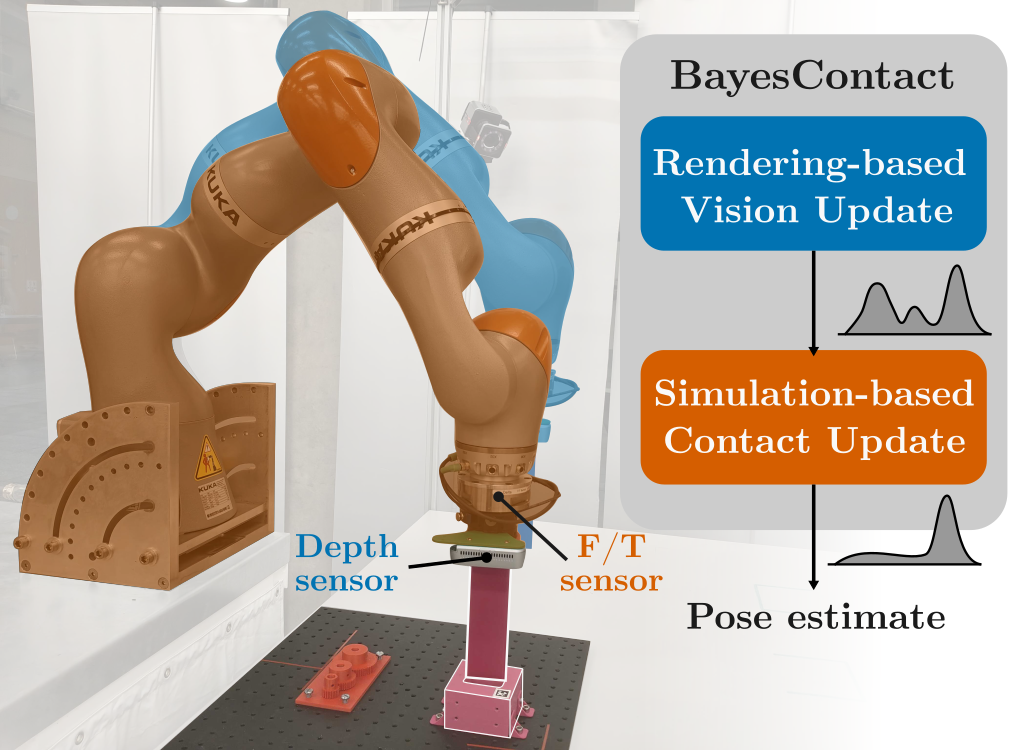}
\caption{\textbf{BayesContact:} We propose a Simulation-based Inference approach to infer uncertain pose using rendering-based depth and physics-based contact likelihoods, which enable online estimation of pose from depth and contact sensing. BayesContact improves pose observability and insertion success over vision-only approaches by
30\%.
}
\end{figure}

Prior work addresses this challenge by training learning-based methods from
simulation via Reinforcement Learning and using Sim2Real techniques to transfer
to real robot settings, or demonstrations via Imitation
Learning~\cite{noseworthy_forge_2025,,luo_precise_2024,
tang_industreal_2023}.
Although effective, these methods require new demonstrations and retraining to achieve insertions in new objects and scenes.
Alternatively, \citet{kim_active_2022} show that decoupling state and dynamics
uncertainty by explicitly estimating pose improves sample efficiency of the
learned policy and Sim2Real transfer.
Motivated by this work, we propose to go one step further and achieve pose estimation for insertion using the same simulation tools used for policy learning but without offline training.



Our key insight is that force--torque (F/T) sensing becomes spatially informative when
interpreted through geometry and simulation. A raw wrench does not directly
reveal a contact point or object pose. However, conditioned on the robot state,
the probing action, and a hypothesized object pose, a physics simulation
can predict what contact evidence should be observed. Comparing simulated and
measured contact therefore defines a likelihood over possible poses. In this
way, contact is not merely a feedback signal for control, but a source of
Bayesian evidence about the hidden scene state. 

To this effect, we present \emph{BayesContact}, an online visuo-contact pose estimation
framework leveraging Simulation-Based Inference (SBI). Inspired by existing works in robotics ~\cite{ramos_bayessim_2019, gothoskar_3dp3_2021,
jatavallabhula_bayesian_2023}, we apply this framework to a contact-rich manipulation
problem.
BayesContact maintains a particle
belief over the pose of a known target object. 
For each pose hypothesis, a graphics simulator predicts visual observations and a physics simulator predicts
contact observations. These simulated observations are compared against real depth and F/T measurements to construct approximate likelihoods, which
are fused through sequential Bayesian inference. We implement this framework using probabilistic programming, allowing rendering-based and
physics-based likelihoods to be composed modularly while supporting multimodal beliefs.




Furthermore, BayesContact also leverages the current belief to choose the next action for probing, turning contact into an active sensing mechanism.

The specific contributions of this work are:
\begin{enumerate}
    \item We introduce a Bayesian Simulation-Based Inference framework~\cite{cranmer_frontier_2020} for online pose estimation with visual and tactile sensing in contact-rich insertion tasks.
    \item We propose rendering-based depth and simulation-based contact likelihoods for flexible estimation.
    \item We demonstrate, through simulation and real robot experiments, that fusing depth vision and F/T contact improves pose observability across multiple insertion geometries.
\end{enumerate}
\section{Related Work}
\subsection{Multi sensor fusion for Contact-rich Manipulation}
Sensor fusion methods integrate diverse information sources into a shared state representation that is suitable for task execution.
In contact-rich manipulation, the sources include cameras, F/T sensing, and robot proprioception.

Visual sensing provides rich task-relevant global information within the context contact-rich manipulation.
However, due to a limited field of view, vision-only approaches are often hindered by partial observability of the task space.
To combat this, some approaches use multiple cameras within the workspace to estimate task-relevant state~\cite{wen_foundationpose_2024}~\cite{lin_sam-6d_2023}.
This is an effective strategy for most contact-rich tasks involving collision free motions.
However, for tasks that involve forceful interactions, multi-camera configurations may not exhaustively cover the task space for enhanced coverage, making contact information vital~\cite{murali_shared_2024}.

Prior works use learned contact process models within a Bayesian Inference framework for effectively fusing contact and vision information to estimate task-relevant state.
\citet{kloss_how_2021} propose differentiable versions of classical filtering methods that learn non-linear process models for contact.
They demonstrate an improved state estimate for a planar-pushing task, over an end-to-end LSTM based learned method.
\citet{lee_making_2019} learn joint sensor observation models for contact and vision using neural networks.
Using a Bayesian Inference framework was shown to improve the estimation of task-relevant state while keeping the interpretability, over end-to-end learned methods.

While these methods are effective for planar pushing tasks, forceful interaction require contact models that relate interacting object configurations with interaction forces. In this work we build upon the Bayesian Inference framework used by prior methods but we integrate sensor likelihoods provided by simulators, particularly graphics and physics engine, which provide additional modeling flexibility.

\subsection{Object Pose Estimation Using Contact}

Prior works attempt to estimate pose of an object using only contact sensing.
\citet{suresh_tactile_2021} jointly estimate pose and geometry of an object in a
planar pushing task using a probabilistic representation and Factor Graph
optimization. They use limit surface models to capture interaction dynamics
with the assumption of known contact location on the robot. These provide
effective contact model approximations that are suitable for the planar pushing
task with the assumption of known contact location. However, forceful
interaction tasks like peg-in-hole insertion require approximations that track
contact location, which are available via simulation when conditioned on
geometry.

\citet{lee_uncertain_2023} propose differentiable contact features to be able to
use contact models for forceful interaction tasks. They demonstrate a
bi-level optimization with the inner optimization solving for contact forces and
the outer optimization solving for object configuration. Our method follows a
similar approach with the exception of using sampling-based inference for
object configuration proposals and a simulator for the contact model evaluation.
This is further demonstrated on a real-world peg insertion task.

Tactile sensing is a common paradigm to estimate and correct for pose
differences. \citet{kim_active_2022} use tactile sensing modalities and
learn a contact line estimator for a peg-in-hole task. This builds upon a factor
graph-based optimization to estimate contact line from tactile sensing.
~\citet{suresh_neural_2023} learn a tactile sensor model to fuse tactile imagery with
visual depth image to jointly estimate pose and reconstruct geometry.

Though effective, these approaches make assumptions of known extrinsic contact
locations. In this work we infer contact locations and fuse the resulting evidence to estimate the hidden hole pose.

\subsection{Sampling-based Inference and Contact}
%
The general Approximate Inference frameworks proposed for the pose estimation tasks involve fusing information via factor graphs. 
When integrating various sensing modalities, factor graph-based optimization require a differentiable forward model to be specified.
Due to the inherent discontinuous and non-linear nature of contact, often task specific contact approximations make them amenable to state inference~\cite{suresh_tactile_2021}~\cite{kim_active_2022}~\cite{lee_uncertain_2023}.

An alternate approach is to use a simulator to specify these forward models.
Simulation-based Inference provides an alternative Approximate Inference that
allow inverting complex forward models~\cite{cranmer_frontier_2020}. Prior works
in robotics have used graphics engines to specify forward models for 3D scene
perception~\cite{gothoskar_3dp3_2021}. This involves
sampling in the parameter space and evaluating simulation-based likelihoods with
observed data to infer unobserved parameters. Often sampling-based approximate
inference methods such as Markov Chain Monte Carlo (MCMC) and Sequential Monte
Carlo (SMC) methods are used to approximate particle-based approximate
distributions of the latent parameter. \citet{jatavallabhula_bayesian_2023} uses simulation-based inference methods to
estimate the parameters of the object articulation model through interaction.

In this work,
we extend this formulation with geometry-conditioned force likelihoods and
depth-based image likelihoods, enabling contact-aware scene understanding for
estimating the hole pose in a peg-insertion task.


%

\section{Background}
\label{sec:background}

\subsection{Simulation-Based Bayesian Inference}
\label{sec:background_sbi}

We formulate pose estimation as Bayesian inference over an unknown object pose \(\mathbf{x}\). Given an observation \(\mathbf{o}\), Bayesian inference updates a prior belief into a posterior,
\begin{equation}
    p(\mathbf{x} \mid \mathbf{o})
    =
    \frac{
        p(\mathbf{o} \mid \mathbf{x})p(\mathbf{x})
    }{
        \int p(\mathbf{o} \mid \mathbf{x}')p(\mathbf{x}')\,d\mathbf{x}'
    }.
    \label{eq:bayes_rule}
\end{equation}
For sequential observations, this update can be written recursively as
\begin{equation}
    b_k(\mathbf{x})
    =
    p(\mathbf{x} \mid \mathbf{o}_{1:k}, \mathbf{a}_{1:k})
    \propto
    p(\mathbf{o}_k \mid \mathbf{x}, \mathbf{a}_k)\, b_{k-1}(\mathbf{x}),
    \label{eq:recursive_bayes_update}
\end{equation}
where \(b_k(\mathbf{x})\) is the belief after the \(k\)-th measurement, \(\mathbf{o}_k\) is the observation, and \(\mathbf{a}_k\) is the action used to acquire it.

In contact-rich manipulation, the likelihood \(p(\mathbf{o}_k \mid \mathbf{x}, \mathbf{a}_k)\) is difficult to specify analytically.
Visual observations depend on rendering, visibility, occlusion, and sensor noise, while contact observations depend on geometry, robot motion, and contact physics. We therefore treat pose estimation as a Simulation-Based Inference problem. For each pose hypothesis \(\mathbf{x}\), a renderer or physics simulator predicts an observation, and a surrogate likelihood scores the agreement between the simulated and measured observations. We implement this generative model using probabilistic programming.

\subsection{Sequential Monte Carlo}
\label{sec:background_smc}

Since the posterior over object pose can be multimodal due to occlusion, symmetry, visual ambiguity, and ambiguous contact explanations, we represent the belief using a weighted set of particles,
\begin{equation}
    b_k(\mathbf{x})
    \approx
    \sum_{i=1}^{N}
    w_k^{(i)}
    \delta\!\left(\mathbf{x} - \mathbf{x}_k^{(i)}\right),
    \label{eq:particle_belief}
\end{equation}
where \(\mathbf{x}_k^{(i)}\) is the \(i\)-th pose hypothesis and \(w_k^{(i)}\) is its normalized weight.

Sequential Monte Carlo (SMC) updates this belief by proposing pose hypotheses, evaluating their simulation-based likelihoods, normalizing particle weights, and resampling when required. This representation is well suited to BayesContact because each particle can be independently rendered or simulated to evaluate its consistency with visual and contact measurements.

\section{Method}
\label{sec:method}

\begin{figure*}[t]
  \centering
  \captionsetup{font=footnotesize}
  \includegraphics[width=\linewidth]{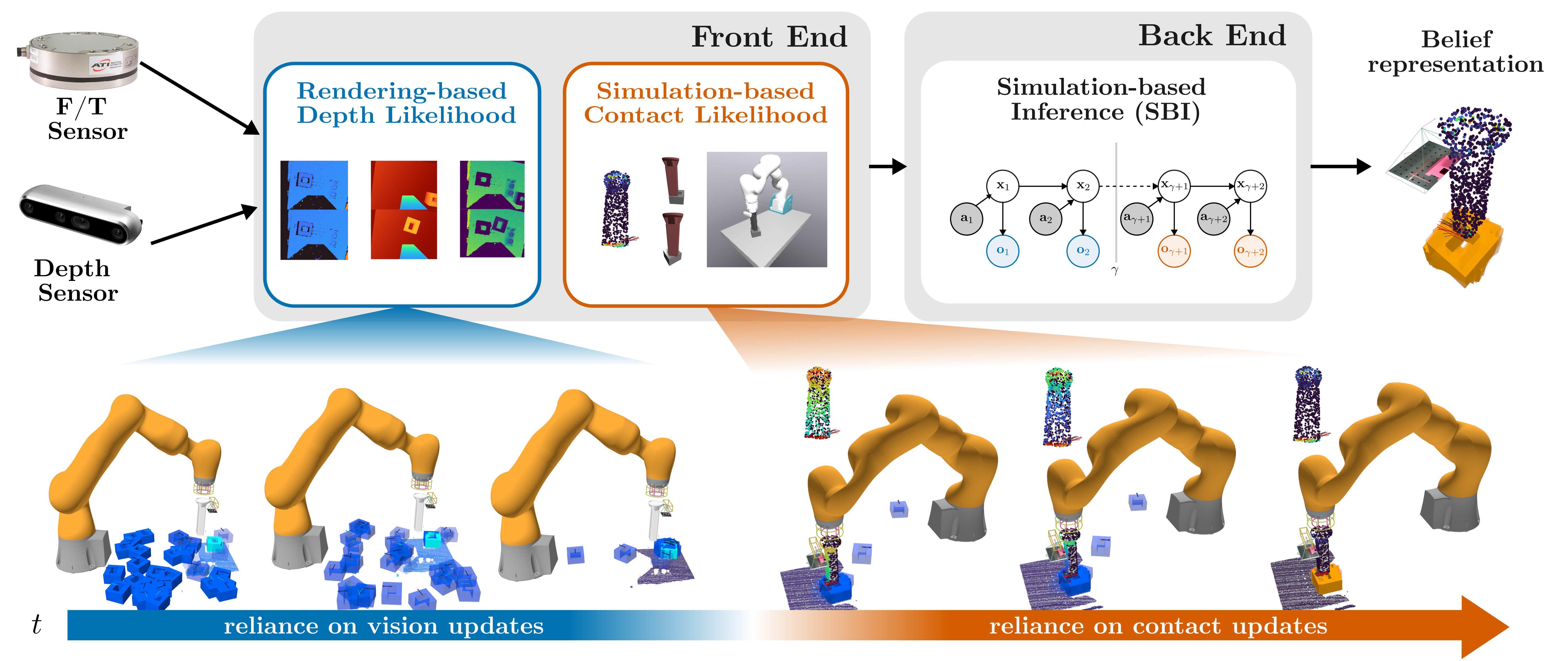}
  \caption{\textbf{System overview.} BayesContact performs pose estimation using simulation-based forward models with a graphics renderer for depth observations and a physics simulator for contact observations. These likelihoods serve as the frontend to a Sequential Monte Carlo inference backend. Together, these allow the robot to estimate belief over pose of a hole in a peg-insertion setup.  
  }
  \label{fig:pp-mc}
\end{figure*}

\begin{figure}
    \centering
    \captionsetup{font=footnotesize}
    \includegraphics[width=1\linewidth]{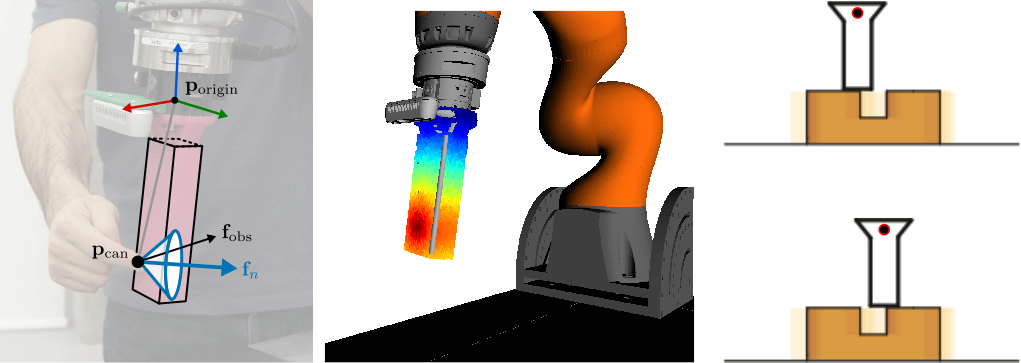}
    \caption{\textbf{Geometry-conditioned Contact Likelihood.} Force-Torque data consists of rich local information about the direction of the hole when conditioned on geometry.}
    \label{fig:f_t_intuition}
\end{figure}

\subsection{Overview}
\label{sec:method_overview}

\textit{BayesContact} estimates the pose of a known \textit{hole} object in a peg-in-hole
task. The state is $\mathbf{x} \in \mathcal{X} \subset SE(3)$, where the pose is parameterized as $\mathbf{x} = (x_x, x_y, \theta).$ Here,
\(x_x\) and \(x_y\) are planar translations, and \(\theta\) is the yaw angle
about the vertical axis. The full object transform is obtained by embedding this
planar pose into \(SE(3)\) with fixed height, roll, and pitch. The pose is estimated using a particle belief, which is updated from depth and F/T measurements. The approach is summarized in \cref{fig:pp-mc}

For inference, each particle is used to determine depth observations from 
a graphics renderer, and contact observations under probing actions from a physics simulator. These simulated observations are compared with the measured depth and F/T data using likelihoods. These determine the front-end of the estimator. The likelihoods are then used to update the belief (via the particles) using SMC inference, which corresponds to the back-end of BayesContact~\cite{carlone2026handbook}.

Inference proceeds in two stages. First, depth observations initialize a vision posterior using SMC with Metropolis--Hastings refinement. Second, the contact phase warm-starts with the vision posterior and updates the belief using contact-rich probing. During this phase, each probing step may provide both F/T-derived contact evidence and associated depth observations, which are fused through a joint simulation-based likelihood.

\subsection{Problem Formulation}
\label{sec:method_problem}

Measurements are indexed by \(k\). The first \(\gamma\) measurements correspond to the visual perception phase. The remaining measurements correspond to contact-rich interaction:
\begin{equation}
    \mathbf{o}_k =
    \begin{cases}
        \mathbf{o}^d_k, & k \leq \gamma,\\
        (\mathbf{o}^f_k,\mathbf{o}^d_k), & k > \gamma.
    \end{cases}
    \label{eq:modality_indexed_observations_method}
\end{equation}
Here, \(\mathbf{o}^d_k\) denotes a depth observation, and \(\mathbf{o}^f_k\)
denotes F/T-derived contact evidence. In the contact phase, the
observation may contain both the contact evidence from the guarded probe and
additional depth information collected during or after the probe. Given action
\(\mathbf{a}_k\), the inference objective is to estimate
\begin{equation}
    b_k(\mathbf{x})
    =
    p(\mathbf{x} \mid \mathbf{o}_{1:k}, \mathbf{a}_{1:k}).
    \label{eq:method_belief}
\end{equation}
The object pose \(\mathbf{x}\) is static across measurements, while the belief \(b_k\) is
updated sequentially as new visual and contact observations are acquired. The upcoming sections discuss the main components of the belief update, namely the observation likelihoods and the SMC inference machinery.

\subsection{Observation Likelihoods}
\label{sec:method_likelihoods}

BayesContact compares observed and simulated measurements using modality-specific likelihoods. These likelihoods are not intended to be exact analytic sensor models. Instead, they act as simulation-based scoring functions: a pose hypothesis receives a high score when its simulated visual or contact evidence is consistent with the measured data. All \(\mathcal{L}\) terms below denote log likelihoods.

\noindent\textbf{Depth Likelihood:}
\label{sec:method_depth_likelihood}
For a visual measurement, the observed depth image \(\mathbf{o}_k^d\) is
compared with a rendered depth image
\(\tilde{\mathbf{o}}_k^d=\mathcal{R}(\mathbf{x})\), where $\mathcal{R}$ is a
function that renders the scene in the world frame with a known camera pose.
The depth likelihood for the current measurement is

\begin{equation}
    \mathcal{L}_{\mathrm{depth}}(\mathbf{o}_k^d \mid \mathbf{x})
    =
    \frac{1}{HW}
    \sum_{u,v}
    \ell_{u,v}(\mathbf{o}_k^d \mid \mathbf{x}),
    \label{eq:depth_likelihood_method}
\end{equation}
where \(H,W\) are the image dimensions. 
$\ell_{u,v}$ denotes a pixel-wise score computed over each pixel $(u,v)$ using a Laplace inlier model mixed with a
constant outlier model:
\begin{equation}
\begin{aligned}
    \ell_{u,v}(\mathbf{o}_k^d \mid \mathbf{x})
    &=
    \log
    \bigg[
    (1-p_{\mathrm{out}})
    {Lap}
    \!\left(
    \mathbf{o}_{k,u,v}^d;
    \tilde{\mathbf{o}}_{k,u,v}^d(x), b
    \right)
    \\
    &\qquad
    +
    p_{\mathrm{out}}\exp(c_{\mathrm{out}})
    \bigg],
\end{aligned}
\label{eq:pixel_depth_likelihood_method}
\end{equation}
where
\[
Lap(\mathbf{y};\mathbf{\mu},b)=\frac{1}{2b}\exp\!\left(-\frac{|\mathbf{y}-\mathbf{\mu}|}{b}\right).
\]
Here $b$ is the scale parameter, \(p_{\mathrm{out}}\) is the outlier probability, and \(c_{\mathrm{out}}\) is the constant outlier log-probability. 

This likelihood assigns high score to pose hypotheses whose rendered depth image
aligns with the observed depth image. The outlier mixture prevents a small
number of mismatched pixels, missing depth returns, or rendering artefacts from
dominating the score.

\vspace{3pt}
\noindent\textbf{Geometry-Conditioned Contact Likelihood:}
\label{sec:method_contact_likelihood}
The contact likelihood compares the observed and simulated contact evidence using a Chamfer-distance score:
\begin{equation}
    \mathcal{L}_{\mathrm{contact}}(\mathbf{o}_k^f \mid \mathbf{x}, \mathbf{a}_k)
    =
    -\sum_{p_i \in \tilde{o}_k^f(x,a_k)}
    \min_{q_j \in o_k^f}
    \left\|
    p_i-q_j
    \right\|^2.
    \label{eq:contact_likelihood_method}
\end{equation}

For contact measurements, the F/T signal is first converted into contact
evidence on the peg surface. Candidate contact points \(\{p_j\}_{j=1}^{P}\) are
sampled from the peg mesh, with corresponding surface normals
\(\{\hat{n}_j\}_{j=1}^{P}\). We assume that the transform from the F/T sensor
frame to the peg frame is known to express the F/T sensor origin relative to the peg. For each candidate point, we use the transform to compute
the moment arm \(\mathbf{r}_j = \mathbf{p}_{\text{can}} - \mathbf{p}_{\text{origin}}\), with the measured force \(\mathbf{f}\) and torque
\(\boldsymbol{\tau}\) expressed in a common frame (\cref{fig:f_t_intuition}).

For a true point contact, we compute the wrench consistency residual
\begin{equation}
    \rho_j
    =
    \left\|
    \boldsymbol{\tau} - (\mathbf{r}_j \times \mathbf{f})
    \right\|_2.
    \label{eq:wrench_residual_method}
\end{equation}
Candidate points that violate simple physical constraints, such as tensile
contact or friction-cone consistency, are down-weighted. The residual
is converted into a score,
\begin{equation}
    s_j = \exp(-\rho_j),
    \label{eq:wrench_score}
\end{equation}
and the highest-scoring candidates define the observed contact evidence \(\mathbf{o}_k^f\).

To obtain the simulated contact measurement, from a pose hypothesis \(\mathbf{x}\) and probe action \(\mathbf{a}_k\), the physics simulator $\mathcal{S}$ predicts the simulated contact evidence
\begin{equation}
    \tilde{\mathbf{o}}_k^f(\mathbf{x},\mathbf{a}_k)
    =
    \mathcal{S}(\mathbf{x},\mathbf{a}_k).
    \label{eq:simulated_contact_evidence_method}
\end{equation}

This likelihood is the main mechanism by which contact improves observability.

\vspace{3pt}
\noindent\textbf{Joint Contact-Phase Likelihood:}
\label{sec:method_joint_likelihood}
During the visual phase, particle weights are updated using only the depth
likelihood. During the contact phase, where ($k>\gamma$), and $\mathbf{o}_k =
(\mathbf{o}_k^f, \mathbf{o}_k^d)$, we employ a joint log-likelihood given by,
\begin{equation}
\begin{aligned}
\mathcal{L}_{\mathrm{joint}}
&(\mathbf{o}_k^f,\mathbf{o}_k^d,\tau_k^{\mathrm{cam}}
\mid \mathbf{x},\mathbf{a}_k)
=
\mathcal{L}_{\mathrm{contact}}(\mathbf{o}_k^f \mid \mathbf{x},\mathbf{a}_k)
\\
&+
\mathcal{L}_{\mathrm{depth}}(\mathbf{o}_k^d \mid \mathbf{x})
+
\mathcal{L}_{\mathrm{traj}}(\tau_k^{\mathrm{cam}} \mid \mathbf{x},\mathbf{a}_k).
\end{aligned}
\label{eq:joint_likelihood_method}
\end{equation}

Where \(\tau_k^{\mathrm{cam}}\) denotes the observed camera trajectory associated with action \(\mathbf{a}_k\). The main contact-driven update comes from \(\mathcal{L}_{\mathrm{contact}}\), which scores whether the simulated contact geometry agrees with the F/T-localized contact evidence.

The trajectory term \(\mathcal{L}_{\mathrm{traj}}\) penalizes mismatch between observed and simulated camera poses using translational and geodesic rotational error. The likelihood is defined as
\begin{equation}
\begin{aligned}
\mathcal{L}_{\mathrm{traj}}
&(\tau_k^{\mathrm{cam}} \mid \mathbf{x},\mathbf{a}_k)
\\
&=
-\sum_{i=1}^{K_{\mathrm{traj}}}
\left[
\frac{
d_t(\mathbf{t}_{k,i}^{\mathrm{obs}},
\mathbf{t}_{k,i}^{\mathrm{sim}})^2
}{
2\sigma_t^2
}
+
\frac{
d_R(\mathbf{R}_{k,i}^{\mathrm{obs}},
\mathbf{R}_{k,i}^{\mathrm{sim}})^2
}{
2\sigma_R^2
}
\right],
\end{aligned}
\label{eq:trajectory_likelihood_method}
\end{equation}
where \( \mathbf{t}_{*,i}^{\mathrm{sim, obs}}, \mathbf{R}_{*,i}^{\mathrm{sim, obs}} \) are simulated and observed translations and orientations of the camera at step $i$ in the probe trajectory. 
The translational and rotational trajectory distances are
\begin{equation}
d_t(\mathbf{t},\hat{\mathbf{t}})
=
\|\mathbf{t}-\hat{\mathbf{t}}\|_2,
d_R(\mathbf{R},\hat{\mathbf{R}})
=
\cos^{-1}
\left(
\frac{\operatorname{tr}(\mathbf{R}\hat{\mathbf{R}}^\top)-1}{2}
\right).
\label{eq:translation_distance}
\end{equation}


\subsection{Sequential Monte Carlo Inference}
\label{sec:method_smc}

BayesContact performs inference in two stages: a vision initialization phase followed by sequential contact updates.

\vspace{3pt}
\noindent\textbf{Vision Phase:}
\label{sec:method_vision_sir_mh}
The vision phase uses Sampling-Importance-Resampling (SIR)~\cite{doucet_tutorial_nodate} with Metropolis--Hastings (MH) rejuvenation to construct an initial posterior from depth observations. At measurement step \(k\), particles (indexed by $j$) are proposed from the explore--exploit proposal
\begin{equation}
    \pi_k(\mathbf{x})
    =
    \alpha \mathcal{U}(\mathcal{X})
    +
    (1-\alpha)
    \sum_j
    w_{k-1}^{(j)}
    \mathcal{K}
    \!\left(
    \mathbf{x}
    \mid
    \mathbf{x}_{k-1}^{(j)}
    \right).
    \label{eq:explore_exploit_proposal}
\end{equation}
Here, \(\mathcal{U}(\mathcal{X})\) is a uniform distribution over the bounded pose-search space, and \(\mathcal{K}(\mathbf{x}\mid\mathbf{x}_{k-1}^{(j)})\) is a Gaussian perturbation kernel centered at particle \(\mathbf{x}_{k-1}^{(j)}\). The coefficient \(\alpha\) is the exploration probability: with probability \(\alpha\), a particle is drawn from \(\mathcal{U}(\mathcal{X})\); otherwise, a previous particle is sampled according to \(w_{k-1}^{(j)}\) and perturbed by \(\mathcal{K}\). For the initial belief \(b_0\), \(\alpha=1\), giving a global uniform proposal over the specified support. Later, \(0<\alpha<1\) balances global exploration and posterior-local exploitation.

Each proposed particle is rendered and scored using the depth log-likelihood. The log importance weight is
\begin{equation}
    \log \tilde{w}_k^{(i)}
    =
    \mathcal{L}_{\mathrm{depth}}
    \!\left(
    \mathbf{o}_k^d
    \mid
    \mathbf{x}_k^{(i)}
    \right)
    +
    \log p_0
    \!\left(
    \mathbf{x}_k^{(i)}
    \right)
    -
    \log \pi_k
    \!\left(
    \mathbf{x}_k^{(i)}
    \right).
    \label{eq:vision_sir_weight}
\end{equation}
The correction \(\log p_0(\mathbf{x})-\log \pi_k(\mathbf{x})\) accounts for sampling from the proposal rather than the prior. For a uniform prior over \(\mathcal{X}\), the prior term is constant within the support.

After the SIR update, particles are refined using random-walk Metropolis--Hastings. The proposal perturbs planar translation and yaw,
\begin{equation}
    \mathbf{x}'
    =
    (x_x+\epsilon_x,\;x_y+\epsilon_y,\;\theta+\epsilon_\theta),
    \label{eq:mh_proposal}
\end{equation}
where \(\epsilon_x,\epsilon_y,\epsilon_\theta\) are zero-mean Gaussian perturbations. With a symmetric proposal and uniform prior, the acceptance probability is
\begin{equation}
\begin{aligned}
    A(\mathbf{x}\rightarrow \mathbf{x}')
    =
    \min
    \big(
    1,
    &\exp
    \big[
    \mathcal{L}_{\mathrm{depth}}
    \!\left(
    \mathbf{o}_k^d
    \mid
    \mathbf{x}'
    \right)
    - \\
    & \mathcal{L}_{\mathrm{depth}}
    \!\left(
    \mathbf{o}_k^d
    \mid
    \mathbf{x}
    \right)
    \big]
    \big).
\end{aligned}
    \label{eq:mh_acceptance}
\end{equation}
The resulting vision posterior initializes the contact phase.

\vspace{3pt}
\noindent\textbf{Contact Phase:}
\label{sec:method_contact_sir}
For contact measurements \(k>\gamma\), an action \(\mathbf{a}_k\) is selected and executed, producing \(\mathbf{o}_k=(\mathbf{o}_k^f,\mathbf{o}_k^d)\), where \(\mathbf{o}_k^f\) is F/T-derived contact evidence and \(\mathbf{o}_k^d\) is the available depth information. Particles are again proposed using \cref{eq:explore_exploit_proposal}, typically with \(\alpha\) close to zero to emphasize exploitation of the current posterior.

For each proposed particle, the physics simulator produces \(\tilde{\mathbf{o}}_k^f(\mathbf{x}_k^{(i)},\mathbf{a}_k)\), and the renderer produces \(\tilde{\mathbf{o}}_k^d(\mathbf{x}_k^{(i)})\) when depth is available. The contact-phase SIR log weight is
\begin{equation}
\begin{aligned}
    \log \tilde{w}_k^{(i)}
    =
    &\mathcal{L}_{\mathrm{joint}}
    \!\left(
    \mathbf{o}_k^f,
    \mathbf{o}_k^d,
    \tau_k^{\mathrm{cam}}
    \mid
    \mathbf{x}_k^{(i)},
    \mathbf{a}_k
    \right)
    \\
    &+
    \log p_0
    \!\left(
    \mathbf{x}_k^{(i)}
    \right)
    -
    \log \pi_k
    \!\left(
    \mathbf{x}_k^{(i)}
    \right).
\end{aligned}
\label{eq:contact_sir_weight}
\end{equation}
As before, the correction term accounts for proposal sampling. When no explicit proposal-density correction is used, particles are weighted directly by the joint log-likelihood.

At each SIR update, unnormalized weights are normalized
and particles are propagated using systematic resampling. Unlike the vision phase, no MH rejuvenation is applied during contact updates because each likelihood evaluation requires physics simulation.

\subsection{Action Space and Guarded Probing}
\label{sec:method_action_space}
To gather contact measurements, the robot executes contact actions $\mathbf{a}_k$, which specify guarded probing motions used to acquire information about the hole pose. In the contact phase, actions are parameterized as planar and yaw offsets,
\begin{equation}
    \mathbf{a}_k = (\Delta x_k,\Delta y_k,\Delta \theta_k),
    \label{eq:probe_action}
\end{equation}
where \(\Delta x_k\) and \(\Delta y_k\) define the planar probing location and \(\Delta \theta_k\) defines the yaw offset of the peg about the vertical axis.

From the commanded offset, the robot executes a guarded vertical dip along the negative \(z\)-axis. The motion is terminated when either a predefined F/T threshold is exceeded or a maximum displacement is reached. Actions are executed using a Cartesian impedance controller that tracks an action-dependent reference trajectory. The F/T measurements acquired during the guarded motion are converted into contact evidence and used in the corresponding Bayesian update.

\subsection{Active Sensing for Next-Best Probe}
\label{sec:method_active_sensing}
Because contact observations are local, their informativeness depends on where the robot probes. We can further use the belief representation provided by BayesContact to select the next probing action from a finite candidate set
\(\mathcal{A}_{\mathrm{dip}} \subset \mathbb{R}^2 \times SO(2)\),
where each action specifies a planar probing offset and yaw offset.

The action selection procedure is shown in Alg.~\ref{alg:myopic_eig}. The
algorithm approximates a one-step information-gain objective using the current
particle belief. An expected posterior entropy is computed given the current
belief. The action that maximizes the information gain between current belief
and predicted posterior is chosen as detailed in \cref{alg:myopic_eig}.



\begin{algorithm}[t]
\captionsetup{font=footnotesize}
\caption{Myopic Approximate Information Gain for Next-Best Probe
}
\label{alg:myopic_eig}
\begin{algorithmic}[1]
\small
\Require Current belief \(b_k = \{(x_k^i,w_k^i)\}_{i=1}^{K}\) truncated and normalized at top-\(K\), candidate action set \(\mathcal{A}_{\mathrm{dip}}\), simulator \(\mathcal{S}\), renderer \(\mathcal{R}\)
\Ensure Selected probing action \(a^{*}\)

\State Compute prior entropy \(H_0 = -\sum_{i=1}^{K}w_k^i \log w_k^i\)

\ForAll{\(a \in \mathcal{A}_{\mathrm{dip}}\)}
    \State Initialise \(\bar{H}(a) \gets 0\)

    \For{\(i = 1\) to \(K\)} \Comment{\textbf{Expected Entropy Loop}}
        \State Simulate contact evidence \(\tilde{o}_{i}^{f} \gets \mathcal{S}(x^i_k,a)\)
        \State Simulate depth evidence \(\tilde{o}_{i}^{d} \gets \mathcal{R}(x^i_k,a)\)
        \For{\(j = 1\) to \(K\)}  \Comment{\textbf{Predict Entropy}}
            \State \(\hat{w}^{(j)}_{k+1} \gets \) Weight update \cref{eq:contact_sir_weight} 
        \EndFor

        

        \State \(H_{k+1}^i \gets -\sum_{j=1}^{K}\hat{w}_{k+1}^{(j)}\log \hat{w}_{k+1}^{(j)}\)
        \State \(\bar{H}(a) \gets \bar{H}(a) + w_k^i H^i_{k+1}\)
    \EndFor

    \State \(\mathrm{IG}(a) \gets H_0 - \bar{H}(a)\)
\EndFor

\State Select \(a^{*} = \arg\max_{a \in \mathcal{A}_{\mathrm{dip}}}\mathrm{IG}(a)\)
\State \Return \(a^{*}\)
\end{algorithmic}
\end{algorithm}

\section{Experiments}

\begin{figure*}
    \centering
    \captionsetup{font=footnotesize}
    \includegraphics[width=1\textwidth]{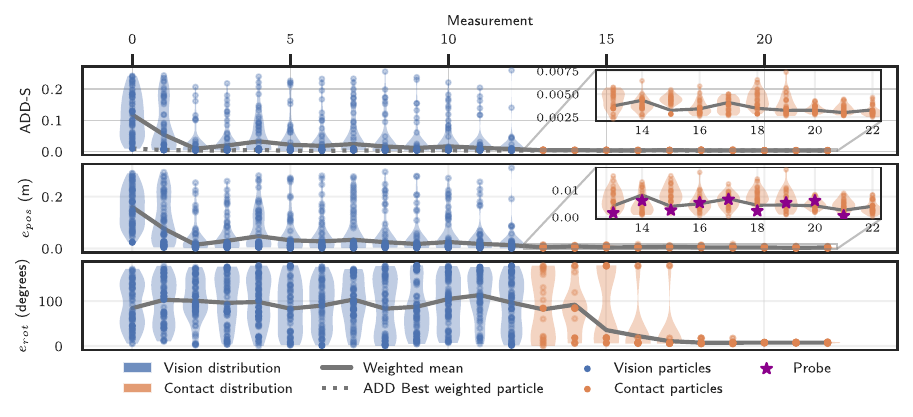}
    \caption{\textbf{Pose convergence.} We illustrate the pose estimate convergence through vision and contact measurements for an example scenario using the \textsc{Ellipse-Teeth} geometry. \textit{Top row:} Convergence of the ADD-S metric across time. \textit{Middle row:} Position error. \textit{Bottom row:} Orientation error. The particle orientation errors show a weight clustering around $\{ 0\degree, 90\degree, 180\degree \}$ coming in from the vision posterior, a consequence of geometric ambiguity. Contact observations then help disambiguate between these modes. 
    }
    \label{fig:add_pos_ori_evol}
\end{figure*}



To validate \textit{BayesContact}, we evaluate three questions: 
\begin{itemize} \item[\textbf{Q1}] Does F/T-induced contact-location information improve pose observability?
\item[\textbf{Q2}] Does a particle-based simulation-inference belief better preserve and resolve geometric ambiguity than parametric alternatives?
\item[\textbf{Q3}] Does this belief enable active sensing?
\end{itemize}

\subsection{Metrics} We evaluate pose accuracy using ADD-S, position error, and orientation error. ADD-S~\cite{wen_foundationpose_2024} measures geometric alignment while accounting for object symmetries:
\begin{equation}
\mathrm{ADD\text{-}S} = \frac{1}{N} \sum_{\mathbf{x}_i \in \mathcal{M}} \min_{\mathbf{x}_j \in \mathcal{M}} \left\| \mathbf{R}\mathbf{x}_i+\mathbf{t} - (\mathbf{R}^{\ast}\mathbf{x}_j+\mathbf{t}^{\ast}) \right\|.
\end{equation}
Where $\mathcal{M}=\{\mathbf{x}_i\}_{i=1}^{N}$ denote points sampled from the object model, and $(\mathbf{R},\mathbf{t})$ and $(\mathbf{R}^{\ast},\mathbf{t}^{\ast})$ are the estimated and ground-truth poses.
The translational error is defined as the Euclidean error $e_{\mathrm{pos}}=d_t(\mathbf{t},\mathbf{t}^{\ast})$, while the rotational error is given by the geodesic distance $e_{\mathrm{rot}}=d_R(\mathbf{R},\mathbf{R}^{\ast})$ (\cref{eq:translation_distance}).

\subsection{Protocol}
\begin{figure}[h]
    \centering
    \captionsetup{font=footnotesize}
    \includegraphics[width=1\linewidth]{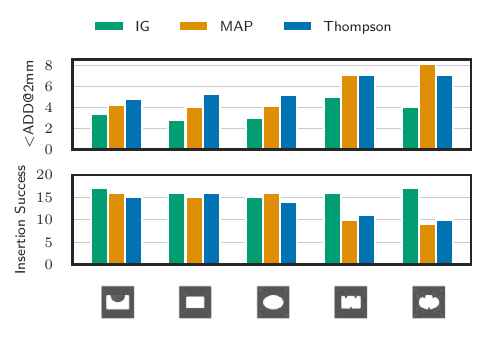}
    \caption{\textbf{Probing efficiency and Insertion Success.} \textit{Top row:} Number of
    contact measurements taken to reach a belief (weighted mean of particles)
    within an ADD of 2 mm, averaged
    across 20 simulated scenarios each. \textit{Bottom row:} Total number of successful insertions out of 20 simulation attempts. Both cases are arranged by the geometry type displayed below: Arch, Rectangle, Ellipse, Rectangle-teeth, and Ellipse-teeth.
    }
    \label{fig:meas_add2}
\end{figure}

We evaluate \textit{BayesContact} (\textbf{BC}) in simulation and on a physical
robot across five peg-in-hole geometries with varying ambiguity
(\cref{fig:meas_add2}). In all trials, the arm is initialized such that the hole
lies within the wrist-camera field of view and is not self-occluded by the robot.

Inference proceeds in two phases. During the vision phase, the robot remains
stationary and updates a pose belief from visual measurements.
Contact phase proceeds with probing strategies as described in the active sensing section below.
Probing terminates when the measurement budget or stopping criterion is reached,
after which the weighted particle mean is used for insertion. 

For each geometry,
we evaluate $20$ randomized object poses in simulation and $10$ in real world.
Inference is performed with $5$ random seeds. We report ADD-S, position error,
and orientation error over measurement index to quantify convergence and
observability.

Simulation experiments and simulation-based
likelihood evaluations are performed in Drake~\cite{tedrake_drake_2019}. The
inference pipeline is implemented in GenJAX. 
The real-world setup uses a 7-DoF
KUKA iiwa14 equipped with an ATI force/torque sensor. Probing actions are
executed with a task-space Cartesian impedance controller using fixed stiffness
and damping gains. Vertical probes are guarded motions along the negative
$z$-axis and terminate when a prescribed force/torque threshold is reached.

\vspace{3pt}
\noindent \textbf{Pose Estimation with vision:} We perform experiments in simulation and the real robot.
We report two
vision-only variants: \textbf{BCv-SMC}, which performs sequential Monte Carlo
inference with MH rejuvenation moves, and \textbf{BCv-PF}, a particle-filter
variant without MH moves. These isolate the quality of the visual posterior
before contact information is introduced. 
We compare against
RANSAC+ICP~\cite{fischler_random_1981,besl_method_1992} (\textbf{ICP}), a
geometric registration baseline, and
FoundationPose~\cite{wen_foundationpose_2024} (\textbf{FP}), a deep learning-based 6D pose estimator.

\vspace{3pt}
\noindent \textbf{Pose Estimation with contact:} During the contact phase, inference is initialized from the vision posterior and
the robot selects probing actions using one of three policies. \textbf{MAP}
probing executes the action associated with the highest-weighted particle.
Thompson Sampling samples a particle according to the current particle weights
and probes according to that hypothesis. Information-gain maximization
(\textbf{IG}) selects the action expected to maximally reduce posterior
uncertainty, as defined in \cref{alg:myopic_eig}. After each guarded probe, the
resulting contact observation is incorporated into the belief. We additionally
compare against a Manifold Unscented Kalman Filter
(\textbf{Man-UKF})~\cite{julier_unscented_2004,murali_artreg_2025}, a parametric
belief-space baseline that fuses vision and contact measurements on the pose
manifold.

\begin{figure}[h]
    \centering
    \captionsetup{font=footnotesize}                             
    \includegraphics[width=1\linewidth]{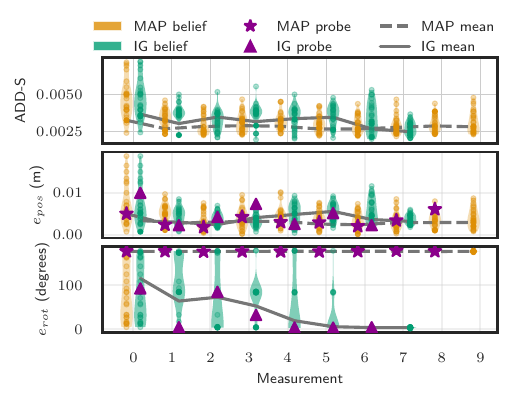}
    \caption{\textbf{Active sensing.} Comparison of \textbf{IG} and the \textbf{MAP} probing strategies. Both strategies are initialized from the same prior vision belief for a single scenario. While they show similar convergence rates on overall ADD-S and position error, \textbf{IG} handles multimodality in orientation better than \textbf{MAP}.
    }
    \label{fig:map_vs_ig}
\end{figure}

\section{Results}

\begin{table*}[h]
  \centering
  \captionsetup{font=footnotesize}
  \caption{\textbf{Summary Pose Estimation Results in Simulation.} We report converged ADD-S (mm), position error(mm), and orientation error (degrees), reported as mean ($\pm$std), for both for Vision-only and Vision + Contact Approaches.
  }
  \resizebox{\textwidth}{!}{
  \begin{tabular}{lllccccccccccccccc}
  \toprule
  & \multirow{2}{*}{\textbf{Method}} 
  & \multicolumn{3}{c}{\textbf{Rectangle}} 
  & \multicolumn{3}{c}{\textbf{Ellipse}} 
  & \multicolumn{3}{c}{\textbf{Rectangle-teeth}} 
  & \multicolumn{3}{c}{\textbf{Ellipse-teeth}} 
  & \multicolumn{3}{c}{\textbf{Arch}}\\
  \cmidrule(lr){3-5} \cmidrule(lr){6-8} \cmidrule(lr){9-11} \cmidrule(lr){12-14} \cmidrule(lr){15-17}
  &
  & ADD-S $\downarrow$ & Pos $\downarrow$ & Ori $\downarrow$ 
  & ADD-S $\downarrow$ & Pos $\downarrow$ & Ori $\downarrow$ 
  & ADD-S $\downarrow$ & Pos $\downarrow$ & Ori $\downarrow$
  & ADD-S $\downarrow$ & Pos $\downarrow$ & Ori $\downarrow$ 
  & ADD-S $\downarrow$ & Pos $\downarrow$ & Ori $\downarrow$\\[2pt]
  \midrule

  \multirow{3}{*}{\rotatebox[origin=c]{90}{\parbox[c]{10mm}{\textbf{Vision + Contact}}}\Tstrut}
  & \textbf{BC-IG}   
  & \shortstack{$\mathbf{3.92}$\\{\scriptsize$\pm1.41$}} 
  & \shortstack{$1.34$\\{\scriptsize$\pm0.31$}} 
  & \shortstack{$34.8$\\{\scriptsize$\pm28.5$}} 
  & \shortstack{$\mathbf{3.74}$\\{\scriptsize$\pm1.36$}} 
  & \shortstack{$1.38$\\{\scriptsize$\pm0.27$}} 
  & \shortstack{$38.1$\\{\scriptsize$\pm31.2$}} 
  & \shortstack{$\mathbf{3.86}$\\{\scriptsize$\pm1.22$}} 
  & \shortstack{$1.21$\\{\scriptsize$\pm0.24$}} 
  & \shortstack{$24.6$\\{\scriptsize$\pm18.9$}} 
  & \shortstack{$\mathbf{3.91}$\\{\scriptsize$\pm1.30$}} 
  & \shortstack{$1.36$\\{\scriptsize$\pm0.29$}} 
  & \shortstack{$27.8$\\{\scriptsize$\pm20.4$}} 
  & \shortstack{$\mathbf{4.12}$\\{\scriptsize$\pm1.47$}} 
  & \shortstack{$1.52$\\{\scriptsize$\pm0.33$}} 
  & \shortstack{$31.7$\\{\scriptsize$\pm24.6$}}\\[2pt]

  & \textbf{BC-MAP}    
  & \shortstack{$4.05$\\{\scriptsize$\pm1.52$}} 
  & \shortstack{$1.42$\\{\scriptsize$\pm0.34$}} 
  & \shortstack{$39.6$\\{\scriptsize$\pm31.7$}} 
  & \shortstack{$3.89$\\{\scriptsize$\pm1.44$}} 
  & \shortstack{$\mathbf{1.36}$\\{\scriptsize$\pm\mathbf{0.30}$}} 
  & \shortstack{$42.5$\\{\scriptsize$\pm34.1$}} 
  & \shortstack{$4.02$\\{\scriptsize$\pm1.28$}} 
  & \shortstack{$1.48$\\{\scriptsize$\pm0.31$}} 
  & \shortstack{$34.9$\\{\scriptsize$\pm25.8$}} 
  & \shortstack{$4.04$\\{\scriptsize$\pm1.37$}} 
  & \shortstack{$1.62$\\{\scriptsize$\pm0.36$}} 
  & \shortstack{$37.6$\\{\scriptsize$\pm27.2$}} 
  & \shortstack{$4.20$\\{\scriptsize$\pm1.51$}} 
  & \shortstack{$1.58$\\{\scriptsize$\pm0.35$}} 
  & \shortstack{$34.8$\\{\scriptsize$\pm26.1$}}\\[2pt]

  & \textbf{Man-UKF}\cite{murali_artreg_2025} 
  & \shortstack{$5.46$\\{\scriptsize$\pm2.08$}} 
  & \shortstack{$1.91$\\{\scriptsize$\pm0.45$}} 
  & \shortstack{$71.2$\\{\scriptsize$\pm49.6$}} 
  & \shortstack{$5.18$\\{\scriptsize$\pm2.01$}} 
  & \shortstack{$1.74$\\{\scriptsize$\pm0.39$}} 
  & \shortstack{$76.8$\\{\scriptsize$\pm52.4$}} 
  & \shortstack{$5.32$\\{\scriptsize$\pm1.86$}} 
  & \shortstack{$1.89$\\{\scriptsize$\pm0.43$}} 
  & \shortstack{$92.5$\\{\scriptsize$\pm47.8$}} 
  & \shortstack{$5.51$\\{\scriptsize$\pm1.94$}} 
  & \shortstack{$2.08$\\{\scriptsize$\pm0.52$}} 
  & \shortstack{$87.9$\\{\scriptsize$\pm50.2$}} 
  & \shortstack{$5.87$\\{\scriptsize$\pm2.15$}} 
  & \shortstack{$2.22$\\{\scriptsize$\pm0.58$}} 
  & \shortstack{$81.4$\\{\scriptsize$\pm46.7$}}\\[2pt]

  \hline
  \multirow{4}{*}{\rotatebox[origin=c]{90}{\parbox[c]{15mm}{\textbf{Vision-only}}}}
  & \textbf{BCv-SMC}  
  & \shortstack{$\mathbf{5.28}$\\{\scriptsize$\pm\mathbf{2.27}$}} 
  & \shortstack{$2.51$\\{\scriptsize$\pm0.46$}} 
  & \shortstack{$76.4$\\{\scriptsize$\pm55.7$}} 
  & \shortstack{$\mathbf{5.11}$\\{\scriptsize$\pm2.39$}} 
  & \shortstack{$2.05$\\{\scriptsize$\pm0.19$}} 
  & \shortstack{$80.4$\\{\scriptsize$\pm58.1$}} 
  & \shortstack{$\mathbf{6.18}$\\{\scriptsize$\pm2.29$}} 
  & \shortstack{$2.07$\\{\scriptsize$\pm0.16$}} 
  & \shortstack{$106.6$\\{\scriptsize$\pm55.1$}} 
  & \shortstack{$\mathbf{6.81}$\\{\scriptsize$\pm4.89$}} 
  & \shortstack{$3.91$\\{\scriptsize$\pm5.16$}} 
  & \shortstack{$83.9$\\{\scriptsize$\pm54.8$}} 
  & \shortstack{$\mathbf{7.08}$\\{\scriptsize$\pm1.40$}} 
  & \shortstack{$2.14$\\{\scriptsize$\pm0.26$}} 
  & \shortstack{$130.4$\\{\scriptsize$\pm40.5$}}\\[2pt]

  & \textbf{BCv-PF}   
  & \shortstack{$6.17$\\{\scriptsize$\pm2.31$}} 
  & \shortstack{$2.66$\\{\scriptsize$\pm0.47$}} 
  & \shortstack{$94.7$\\{\scriptsize$\pm60.5$}} 
  & \shortstack{$5.62$\\{\scriptsize$\pm2.65$}} 
  & \shortstack{$2.24$\\{\scriptsize$\pm0.36$}} 
  & \shortstack{$87.6$\\{\scriptsize$\pm62.2$}} 
  & \shortstack{$6.37$\\{\scriptsize$\pm1.18$}} 
  & \shortstack{$2.39$\\{\scriptsize$\pm0.26$}} 
  & \shortstack{$98.8$\\{\scriptsize$\pm26.1$}} 
  & \shortstack{$6.21$\\{\scriptsize$\pm0.60$}} 
  & \shortstack{$2.46$\\{\scriptsize$\pm0.36$}} 
  & \shortstack{$92.5$\\{\scriptsize$\pm13.7$}} 
  & \shortstack{$13.26$\\{\scriptsize$\pm7.18$}} 
  & \shortstack{$12.02$\\{\scriptsize$\pm8.10$}} 
  & \shortstack{$72.4$\\{\scriptsize$\pm18.2$}}\\[2pt]

  & \textbf{ICP}~\cite{besl_method_1992} 
  & \shortstack{$46.15$\\{\scriptsize$\pm23.47$}} 
  & \shortstack{$14.90$\\{\scriptsize$\pm23.61$}} 
  & \shortstack{$77.8$\\{\scriptsize$\pm63.7$}} 
  & \shortstack{$40.87$\\{\scriptsize$\pm20.22$}} 
  & \shortstack{$13.96$\\{\scriptsize$\pm20.27$}} 
  & \shortstack{$91.3$\\{\scriptsize$\pm65.9$}} 
  & \shortstack{$53.53$\\{\scriptsize$\pm11.64$}} 
  & \shortstack{$18.80$\\{\scriptsize$\pm12.06$}} 
  & \shortstack{$118.1$\\{\scriptsize$\pm51.1$}} 
  & \shortstack{$55.31$\\{\scriptsize$\pm17.88$}} 
  & \shortstack{$14.79$\\{\scriptsize$\pm17.87$}} 
  & \shortstack{$82.1$\\{\scriptsize$\pm59.6$}} 
  & \shortstack{$45.66$\\{\scriptsize$\pm16.91$}} 
  & \shortstack{$15.10$\\{\scriptsize$\pm17.46$}} 
  & \shortstack{$93.3$\\{\scriptsize$\pm51.1$}}\\[2pt]

  & \textbf{FP}~\cite{wen_foundationpose_2024} 
  & \shortstack{$10.85$\\{\scriptsize$\pm6.76$}} 
  & \shortstack{$9.31$\\{\scriptsize$\pm7.93$}} 
  & \shortstack{$102.3$\\{\scriptsize$\pm81.9$}} 
  & \shortstack{$10.69$\\{\scriptsize$\pm4.79$}} 
  & \shortstack{$7.69$\\{\scriptsize$\pm6.96$}} 
  & \shortstack{$142.9$\\{\scriptsize$\pm57.6$}} 
  & \shortstack{$8.17$\\{\scriptsize$\pm3.67$}} 
  & \shortstack{$5.32$\\{\scriptsize$\pm4.07$}} 
  & \shortstack{$100.7$\\{\scriptsize$\pm75.7$}} 
  & \shortstack{$11.08$\\{\scriptsize$\pm6.41$}} 
  & \shortstack{$9.60$\\{\scriptsize$\pm9.01$}} 
  & \shortstack{$110.6$\\{\scriptsize$\pm79.3$}} 
  & \shortstack{$9.70$\\{\scriptsize$\pm5.10$}} 
  & \shortstack{$6.59$\\{\scriptsize$\pm5.71$}} 
  & \shortstack{$107.4$\\{\scriptsize$\pm76.0$}}\\[2pt]

  \bottomrule
  \end{tabular}
  }
  \label{tab:add_metric_results}
\end{table*}

\begin{table}[t]
  \centering
  \captionsetup{font=footnotesize}
  \caption{\textbf{Real-robot results.} We report converged ADD-S error and insertion success rate for Rectangle and Rectangle-teeth.}
  \resizebox{\columnwidth}{!}{
  \begin{tabular}{lcccc}
  \toprule
  \multirow{2}{*}{\textbf{Method}} 
  & \multicolumn{2}{c}{\textbf{Rectangle}} 
  & \multicolumn{2}{c}{\textbf{Rectangle-teeth}} \\
  \cmidrule(lr){2-3} \cmidrule(lr){4-5}
  & ADD-S (mm) $\downarrow$ & Success $\uparrow$
  & ADD-S (mm) $\downarrow$ & Success $\uparrow$ \\
  \midrule
%

  \textbf{BCv-SMC}
  & \shortstack{$6.12$\\{\scriptsize$\pm2.44$}}
  & $3/10$
  & \shortstack{$6.58$\\{\scriptsize$\pm2.71$}}
  & $3/10$ \\

  \textbf{BC-IG}
  & \shortstack{$4.04$\\{\scriptsize$\pm1.94$}}
  & $5/10$
  & \shortstack{$4.21$\\{\scriptsize$\pm1.56$}}
  & $6/10$ \\
  \bottomrule
  \end{tabular}
  }
  \label{tab:real_world_results}
\end{table}

\vspace{3pt}
\noindent\textbf{Contact-Location Observations Improve Pose Observability:} Our experiments in simulation demonstrate that incorporating F/T-induced contact-location information consistently improves pose observability over vision-only inference. As shown in Table~\ref{tab:add_metric_results}, BC-IG reduces average converged ADD-S, position error, and orientation error over BCv-SMC by 35.8\%, 46.3\%, and 67.1\%, respectively, across all simulated geometries. The gains are largest for visually ambiguous geometries: on Rectangle-teeth, BC-IG reduces ADD-S, position error, and orientation error by 37.5\%, 41.5\%, and 76.9\%, respectively; on Ellipse-teeth, the corresponding reductions are 42.6\%, 65.2\%, and 66.9\%. Fig.~\ref{fig:add_pos_ori_evol} illustrates this effect: the vision posterior reduces translational uncertainty but retains multiple orientation modes, while contact observations downweight hypotheses inconsistent with the measured contact evidence. Real-robot results in Table~\ref{tab:real_world_results} show the same trend, with BC-IG reducing ADD-S by 34.0\% on Rectangle and 36.0\% on Rectangle-teeth over BCv-SMC, while improving insertion success by 20\% and 30\%, respectively.

\vspace{3pt}
\noindent\textbf{Particle Beliefs Preserve and Resolve Geometric Ambiguity:}
The particle-based belief representation is important for preserving ambiguity until contact provides disambiguating evidence. Table~\ref{tab:add_metric_results} shows that BC-IG outperforms Man-UKF across all geometries, reducing average ADD-S, position error, and orientation error by 28.5\%, 30.8\%, and 61.7\%, respectively. The largest improvements are again in orientation, where a unimodal parametric belief struggles to represent separated pose modes. On Rectangle-teeth, BC-IG reduces orientation error over Man-UKF by 73.4\%; on Ellipse-teeth, it reduces orientation error by 68.4\%. These results indicate that BayesContact benefits not only from contact sensing, but also from maintaining a multimodal particle belief that can be reweighted as new contact evidence arrives.

\vspace{3pt}
\noindent\textbf{Particle Contact Beliefs Enable Information-Gain Active Sensing:}
Combining contact observations with a particle belief enables active probing. Fig.~\ref{fig:map_vs_ig} shows that both MAP and IG probing improve pose accuracy from the same vision posterior, but IG better targets orientation ambiguity by selecting actions that reduce expected posterior entropy rather than committing to the current maximum-weight hypothesis. Fig.~\ref{fig:meas_add2} suggests that IG reaches the ADD-S threshold with fewer probes and achieves higher insertion success than MAP and Thompson sampling in several geometries. Thus, the current results support contact-driven improvement most strongly, while showing that information-gain probing is a promising mechanism for using the particle belief to actively resolve residual ambiguity.

\section{Conclusion}

In this work we proposed \textit{BayesContact}, a Simulation-Based Inference
framework that incorporates geometry-conditioned F/T likelihoods and depth
likelihoods with SMC inference to estimate the belief over pose of a hole in a
peg-insertion setting. Through simulation and real robot experiments we
demonstrate that F/T based contact improves pose observability over
depth-only estimates. We further demonstrate how the resulting belief
representation can be used for Information Gain-based active sensing.



\printbibliography 
\end{document}